\documentclass{article}

\usepackage{arxiv}
\usepackage[utf8]{inputenc}
\usepackage[T1]{fontenc}
\usepackage{hyperref}
\usepackage{url}
\usepackage{graphicx}
\usepackage{booktabs}
\usepackage{amsmath,amssymb,amsfonts}
\usepackage{xcolor}
\usepackage{listings}
\usepackage{natbib}
\usepackage{caption}

\graphicspath{{./}}

\def\rashomon{\texttt{\textsc{Rashomon Set}}}

\def\credit-approval{\texttt{\textsc{CREDIT-APPROVAL}}}

\def\folktables{\texttt{\textsc{Folktables}}}

\lstdefinestyle{mystyle}{
    basicstyle=\footnotesize\ttfamily,
    breakatwhitespace=false,
    breaklines=true,
    captionpos=b,
    keepspaces=true,
    numbers=left,
    numberstyle=\scriptsize\ttfamily,
    numbersep=5pt,
    showspaces=false,
    showstringspaces=false,
    showtabs=false,
    tabsize=2
}

\def\target{t}
\def\surrogate{s}
\def\inputspace{\mathcal{X}}
\def\outputspace{\mathcal{Y}}

\def\cm{\mathcal{H}}

\def\sp{\Delta_{\operatorname{SP}}}
\def\pe{\Delta_{\operatorname{PE}}}
\def\eop{\Delta_{\operatorname{EOpp}}}
\def\eodds{\Delta_{\operatorname{EOdds}}}

\title{Model Stealing Through the Lens of Model Multiplicity}
\usepackage{authblk}

\author[$\S$]{Eliott Baltz}
\author[$\ddagger$]{Satoshi Hara}
\author[$\S$]{Ulrich A\"ivodji}
\affil[$\ddagger$]{The University of Electro-Communications : Tokyo}
\affil[$\S$]{\'ETS, Mila : Montr\'eal}
\date{}

\begin{document}

\maketitle

\begin{abstract}
Model stealing attacks, where adversaries create high-fidelity surrogate models, are a significant threat to the intellectual property of machine learning services. Conventional wisdom suggests these surrogates could provide adversaries with economic leverage comparable to the original service providers. This paper challenges this assumption by evaluating model stealing attacks beyond mere fidelity to the target model. Because query-based extraction provides only partial supervision of the target's input-output behavior, the surrogate is not uniquely identified: many near-optimal surrogates can achieve comparable fidelity while differing in deployment-relevant properties. Instead of performing a classic learning-based model stealing attack, we compute the Rashomon Set (i.e., the set of almost-equally-accurate models) of surrogate models, and evaluate its diversity using multiplicity metrics (ambiguity, discrepancy, and Rashomon Capacity) and group fairness metrics. Across tabular, medical imaging, and NLP tasks, our experiments on real-world datasets reveal that despite exhibiting similar fidelity to the target model, surrogate models can display significant variances in other critical performance metrics. These findings cast doubt on the presumed equivalence between high-fidelity surrogates and the target model in practical deployment scenarios.
\end{abstract}
\section{Introduction}
Machine learning models have become integral to various applications, yet they remain vulnerable to a range of attacks that threaten their integrity and the business models built around them~\citep{shokri2017membership,fredrikson2014privacy,ateniese2015hacking,carlini2019secret,tramer2016stealing,kumar2020adversarial}. Among these, model stealing attacks~\citep{tramer2016stealing} pose a significant challenge, particularly in the context of Machine Learning as a Service (\texttt{MLaaS}). Adversaries exploit query access to a target model to create a surrogate model that mimics its performance, potentially compromising the intellectual property of the original service provider.
 \begin{figure*}[h!]
    \begin{center}
    \includegraphics[width=0.99\textwidth]{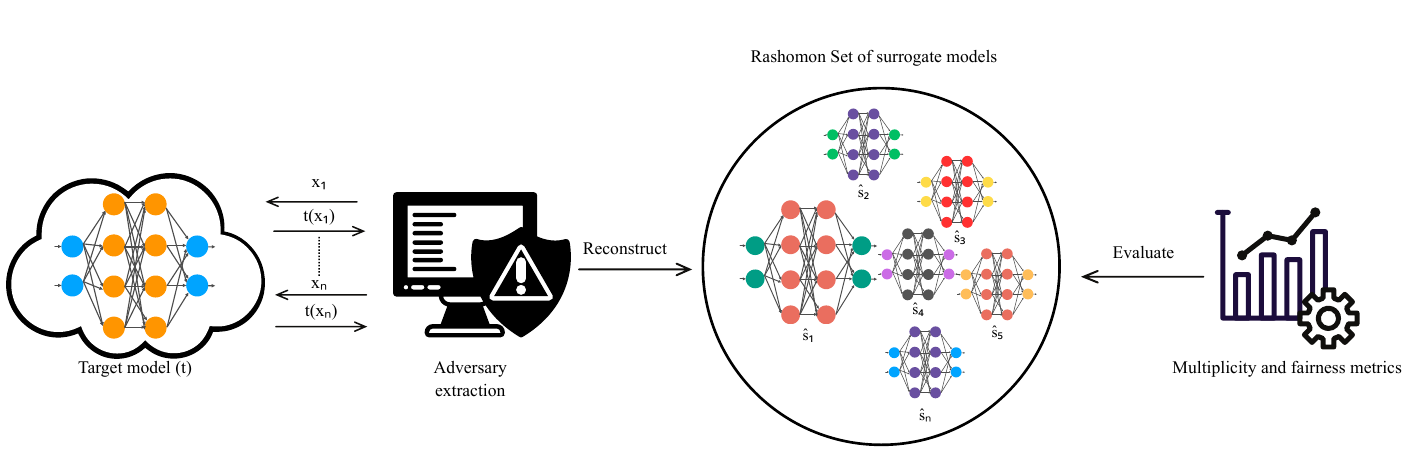}
    \end{center}
    \caption{Overview of extraction-induced multiplicity. An adversary queries the target model $t$ on auxiliary inputs and uses the returned predictions to reconstruct a surrogate model $s$. Rather than treating this surrogate as a unique clone of the target, we study the surrounding \rashomon{} of near-equivalent surrogate models that can arise from the extraction process. We then evaluate this set using fidelity, multiplicity metrics, and group fairness metrics.}
\label{fig:workflow}
\end{figure*}

Traditionally, the success of model stealing attacks has been gauged by the fidelity of the surrogate model, defined as its label agreement with the target model on a reference dataset. High-fidelity surrogates have therefore been interpreted as evidence of a significant intellectual-property loss, suggesting that the adversary has effectively replicated the target model's capabilities. However, recent advancements in our understanding of model multiplicity, particularly through the lens of Rashomon Sets~\citep{semenova2022existence}, challenge this simplistic interpretation.

The Rashomon Set, consisting of models that perform approximately as well as a reference model, has been shown to include models with different practical performance profiles. This diversity manifests in conflicting predictions on individual samples~\citep{marx2020predictive} and varying performance under distribution shifts~\citep{d2022underspecification}. Similarly, research on post-hoc explanations has revealed that high-fidelity explainers, which can be viewed as surrogate models of a different function class, often display disparate performance in fairness metrics, facilitating explanation manipulation attacks such as fairwashing~\citep{aivodji2021characterizing}.

In light of these findings, we pose a critical question: \textbf{Does high-fidelity model stealing truly produce surrogate models with comparable deployment behavior?} Our investigation suggests that claims of intellectual property loss based only on fidelity may be incomplete, and that new metrics are necessary to characterize the similarity between high-fidelity surrogates and their target models. Given the existence of a sizable \rashomon{}, a firm attempting to ``steal'' another firm's model may recover a model that matches overall fidelity yet disagrees with the source model at the instance level. In those contexts, ``successful'' theft judged by fidelity alone can fail to replicate downstream behavior, redistribute errors across populations, and alter individual recourse, even when fidelity is high. Accordingly, evaluating surrogates demands metrics beyond accuracy and fidelity, including ambiguity, discrepancy, Rashomon Capacity, and fairness metrics, since fidelity is an incomplete proxy for functional equivalence.

Viewed through this lens, model stealing is an identification problem under multiplicity: query access constrains the surrogate only on a finite region of the input space, leaving a potentially large set of near-equivalent predictors consistent with the observed query-response pairs. Our evaluation, therefore, targets the structure of this extraction-induced \rashomon{} rather than treating a single high-fidelity surrogate as representative. In this paper, we focus on fidelity extraction. Figure~\ref{fig:workflow} summarizes this setting and highlights the step studied in this work: after reconstructing a surrogate, we evaluate the surrounding extraction-induced \rashomon{} rather than treating the stolen model as a unique clone of the target. We study this phenomenon across three domains: tabular prediction tasks, medical image classification, and NLP sentiment classification. This lets us test whether extraction-induced multiplicity is specific to one model family or whether it persists across neural networks, convolutional architectures, and transformer-based language models.

This paper makes the following key contributions:

\begin{itemize}
    \item We introduce a method for generating dropout-based \rashomon{} sets around extracted surrogate models, allowing us to study model stealing as a multiplicity problem rather than as a single-model fidelity problem.
    \item We evaluate extraction-induced multiplicity across tabular data, medical imaging, and NLP tasks.
    %, using both Knockoff Nets and prediction-API model stealing.
    \item We characterize the diversity of extracted \rashomon{} sets using ambiguity, discrepancy, Rashomon Capacity, and group fairness metrics.
    \item Our empirical results demonstrate that surrogate models can exhibit high fidelity while still displaying different behavior, including conflicting sample-level predictions, attack-dependent multiplicity, and variable fairness values on tabular tasks.
\end{itemize}

These findings cast doubt on the prevailing interpretation of high-fidelity model extraction attacks and their implications for model exploitation. By highlighting the nuanced relationship between fidelity and practical performance across multiple domains, our work paves the way for more comprehensive evaluations of model stealing attacks and their potential impact on \texttt{MLaaS}.

\section{Background \& Related work}
We focus on classification tasks. Let $X \in \inputspace{} \subset \mathbb{R}^n$ denote an input, $Y \in \outputspace{}$ its label, and $\target{} : \inputspace{} \to \outputspace{}$ the target model from a model class $\cm{}$.

\subsection{Multiplicity}

Multiplicity can arise at several stages of the machine learning pipeline. At the \textbf{dataset and preprocessing} level, multiple plausible datasets or preprocessing choices may represent the same phenomenon due to noise, missingness, measurement error, or structural bias, leading to different sets of near-optimal predictors~\citep{meyer2023dataset}. At the \textbf{training} level, predictive multiplicity occurs when several near-optimal models achieve similar performance while disagreeing on individual predictions~\citep{marx2020predictive}. Related work on underspecification shows that different training runs can converge to models with similar in-distribution performance but different downstream behavior~\citep{d2022underspecification}, while representation-level perturbations can also generate diverse families of near-optimal predictors~\citep{eerlings2026diverse}. At the \textbf{explanation} level, a single predictive model may admit multiple explanations that differ substantially in feature attributions or fairness narratives~\citep{aivodji2021characterizing,hwang2026explanation}.

In standard learning pipelines, multiplicity reflects ambiguity in the data, preprocessing, or optimization process. In our setting, multiplicity is induced by the extraction process itself. Because query-based supervision provides only partial information about the target model, the stolen surrogate is identified only up to the constraints imposed by the queried examples. Model stealing, therefore, gives rise to an extraction-induced \rashomon{}: a family of surrogates that exhibit high label agreement with the target yet may diverge elsewhere in individual predictions, fairness, or robustness.

\subsection{Characterization of multiplicity}

We focus on three multiplicity metrics: ambiguity, discrepancy, and Rashomon Capacity. Ambiguity measures how often models in the \rashomon{} disagree on individual predictions; discrepancy measures the worst-case disagreement with a reference model; and Rashomon Capacity measures the dispersion of predictive scores across the set. Formal definitions are given in Section~\ref{sec:multiplicity}.

\subsection{Model stealing attacks}

In model stealing attacks~\citep{tramer2016stealing}, an adversary uses black-box access to a target model $\target{}$ to train a surrogate model $\surrogate{}$. The attacker first queries $\target{}$ on an auxiliary dataset, collects its outputs, and then trains $\surrogate{}$ on these query-response pairs. Since the attack set depends on the query strategy, sampling distribution, and budget, model stealing can be viewed as a data acquisition pipeline whose outputs may induce different surrogate solution sets even for the same target model. Model stealing attacks are commonly divided into task-accuracy extraction, where the surrogate matches the target's task performance, and fidelity extraction, where it reproduces the target's decision behavior~\citep{jagielski2020high}. The fidelity of a surrogate model $\surrogate{}$ with respect to a target model $\target{}$ on a reference set $X_{ref}$ is
\begin{align*}
\mathsf{fidelity}(\surrogate{}) = \frac{1}{|X_{ref}|} \sum_{x \in X_{ref}} \mathbb{I}(\surrogate{}(x)=\target{}(x)).
\end{align*}

\subsection{\rashomon{} and Predictive Multiplicity}

The \emph{Rashomon Effect} refers to the existence of multiple models that explain the same data equally well~\citep{breiman2001statistical}. In machine learning, this motivates the study of the \rashomon{}, the set $S_{\epsilon}(h_0)$ of all models whose empirical risk lies within $\epsilon$ of that of a baseline model $h_0$~\citep{marx2020predictive}:
\begin{equation}
    S_{\epsilon}(h_0) := \{ h \in H : \hat{R}(h) \leq \hat{R}(h_0) + \epsilon \}.
\end{equation}
Predictive multiplicity exists when some $h \in S_\epsilon(h_0)$ disagrees with $h_0$ on at least one input~\citep{marx2020predictive}.

\subsection{Approximation of the \rashomon{}}

Enumerating the full \rashomon{} is infeasible for complex model classes such as multilayer perceptrons. A practical alternative is to use \texttt{Dropout} at inference time to sample models likely to belong to the \rashomon{}~\citep{hsu2024dropout}. Compared with repeated retraining~\citep{breiman2001statistical} or adversarial weight perturbation~\citep{hsu2022rashomon}, dropout-based sampling is much cheaper computationally and allows us to generate large sets of near-equivalent models efficiently.

\subsection{Related work}

To the best of our knowledge, this work is the first to study the fragility of fidelity in model reconstruction attacks through the lens of multiplicity. Prior work has shown that models within a \rashomon{} can differ in fairness or interpretability even when accuracy is similar~\citep{blackModelMultiplicityOpportunities2022}, but this question has not been examined in the context of model stealing. More broadly, the limits of fidelity have already been noted in interpretable machine learning, where high-fidelity explanations can still be misleading or strategically manipulated, for example, in fairwashing settings~\citep{aivodji2019fairwashing,lakkaraju2020fool}. Our work extends this concern to model extraction by asking whether high-fidelity stolen models are genuinely behaviorally equivalent to their targets.

\section{Evaluation metrics}
We evaluate the implications of model stealing attacks using multiplicity and fairness.
 \begin{figure*}[h!]
    \begin{center}
    \includegraphics[width=0.75\textwidth]{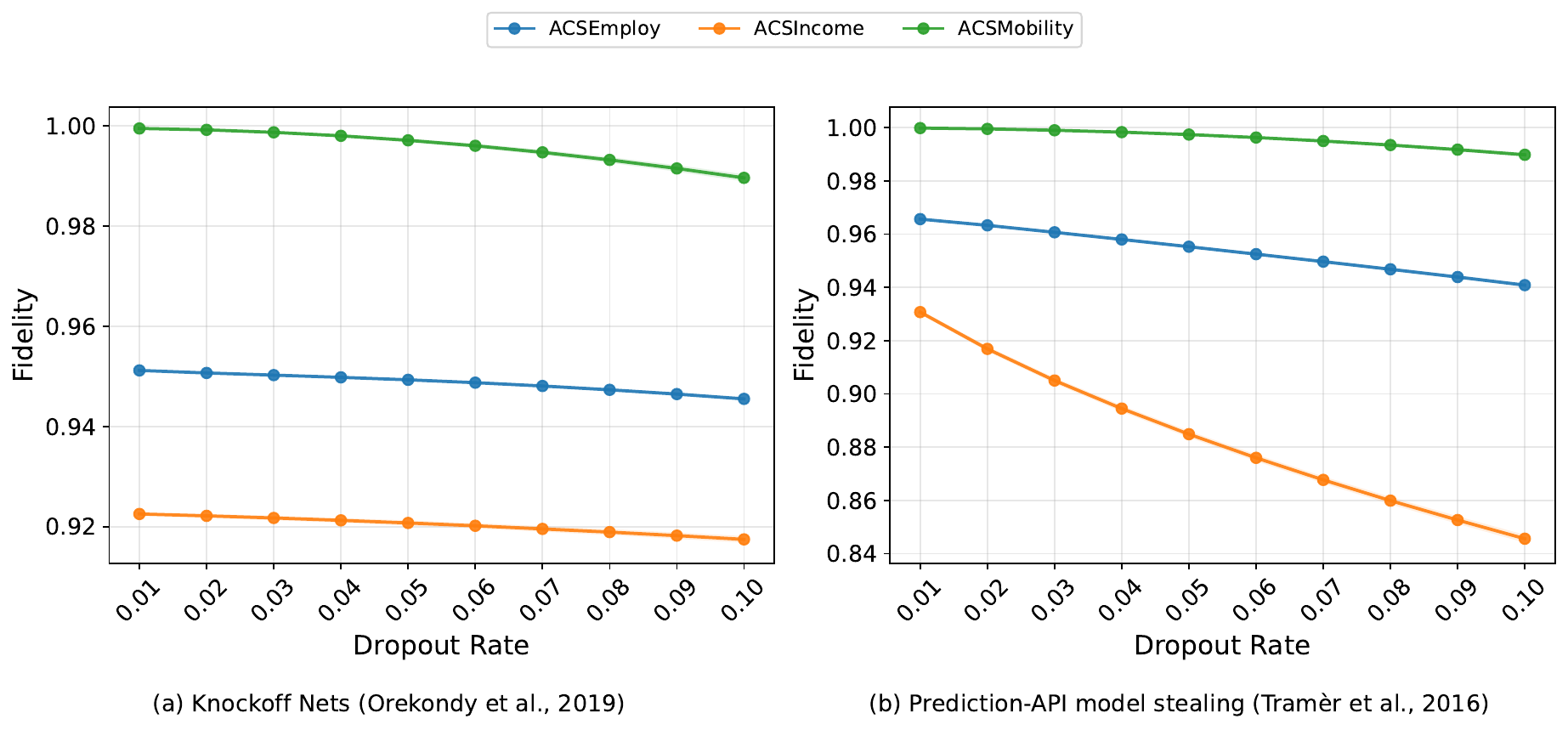}
    \end{center}
    \caption{Fidelity for \rashomon{} of size 5000 for each dropout rate for both attacks on \folktables{} datasets.}
    \label{fig:fid}
\end{figure*}

\subsection{Multiplicity}
\label{sec:multiplicity}

We quantify predictive multiplicity within the \rashomon{} using both label-based and score-based metrics. The notions of discrepancy and ambiguity are introduced to capture conflicting predictions within the \rashomon{} \citep{marx2020predictive}. More precisely, given a \rashomon{} $S_{\epsilon}(h_0)$, the corresponding discrepancy $\delta_{\epsilon}\left(h_{0}\right)$ and ambiguity $\alpha_{\epsilon}\left(h_{0}\right)$ are defined as follows:

\begin{equation}
\delta_{\epsilon}\left(h_{0}\right) := \max_{h \in S_{\epsilon}\left(h_{0}\right)} \frac{1}{n} \sum_{i=1}^{n} \mathbf{1}\left[h\left(\boldsymbol{x}_{i}\right) \neq h_{0}\left(\boldsymbol{x}_{i}\right)\right]
\end{equation}

\begin{equation}
\alpha_{\epsilon}\left(h_{0}\right) := \frac{1}{n} \sum_{i=1}^{n} \max_{h \in S_{\epsilon}\left(h_{0}\right)} \mathbf{1}\left[h\left(\boldsymbol{x}_{i}\right) \neq h_{0}\left(\boldsymbol{x}_{i}\right)\right]
\end{equation}

Discrepancy quantifies the worst-case disagreement rate between the reference model $h_0$ and any other model within the \rashomon{}, while ambiguity captures the average pointwise disagreement between $h_0$ and the \rashomon{}. On one hand, high ambiguity means that across many input instances, there is at least some disagreement between $h_0$ and other models in the Rashomon set, whereas low ambiguity reflects broad consensus (\emph{i.e.,} most models tend to agree with $h_0$ on most data points). On the other hand, a high discrepancy indicates the existence of at least one model whose predictions diverge significantly from those of $h_0$, suggesting potential extremes in model behavior. In contrast, a low discrepancy implies that all competing models make predictions that closely align with $h_0$. These complementary metrics characterize how impactful the choice of model can be for individuals. In high-stakes domains, this shows that even if a model performs well in terms of accuracy, other models may exist that perform equally well while producing substantially different outcomes for some people. This means that an attacker may achieve high-fidelity extraction without reproducing the actual decision rule of the target model.

Beyond label-based disagreement, we also evaluate score-level multiplicity through Rashomon Capacity, which measures how widely predictive probabilities can vary across models in the Rashomon set while maintaining near-optimal performance \citep{hsu2022rashomon}. For a sample $x_i$, the Rashomon Capacity is defined as
\[
m_C(x_i) := 2^{C(M_\epsilon(x_i))},
\]
where $\epsilon \geq 0$ is the Rashomon parameter, $M_\epsilon(x_i)$ denotes the $\epsilon$-multiplicity set of $x_i$ (that is, the set of predictions for $x_i$ induced by models in the \rashomon), and $C(M_\epsilon(x_i))$ is the capacity of that multiplicity set. Intuitively, larger values of Rashomon Capacity indicate greater dispersion of predictive probabilities across near-optimal models, even when these models exhibit similar overall performance. In our setting, this metric complements ambiguity and discrepancy by capturing variability in confidence scores rather than only variability in predicted labels.

\subsection{Fairness}
For fairness evaluation, we consider the case of binary classifiers (\emph{i.e.,} $Y = \{0,1\}$) and focus on statistical notions of fairness. These metrics require a model to exhibit approximate parity for some statistical measure across the different demographic groups defined by a sensitive attribute $G$ (\emph{e.g.,} age, sex). In particular, we assume binary sensitive attribute (\emph{i.e.,} $G = \{0,1\}$) for simplicity and consider four metrics, namely statistical parity ($\sp{}$)~\citep{dwork2012fairness}, predictive equality ($\pe{}$)~\citep{corbett2017algorithmic}, equal opportunity ($\eop{}$) and equalized odds ($\eodds{}$)~\citep{hardt2016equality}. Table~\ref{tab:fairness} presents a summary of the definition of these metrics.

\begin{table}[h!]
\centering
\caption{Summary of the different statistical notions of fairness considered.}
\label{tab:fairness}
\resizebox{0.45\textwidth}{!}{%
\small
\begin{tabular}{ll}
\toprule
\textbf{Metric} & \textbf{Definition} \\
\midrule
$\sp{}$ & $|P(\hat{Y}=1|G=0) - P(\hat{Y}=1|G=1)|$ \\
$\pe{}$ & $|P(\hat{Y}=1|Y=0,G=0) - P(\hat{Y}=1|Y=0,G=1)|$ \\
$\eop{}$ & $|P(\hat{Y}=1|Y=1,G=0) - P(\hat{Y}=1|Y=1,G=1)|$ \\
\shortstack{\vspace{6pt}$\eodds{}$} 
& 
\shortstack[l]{
$|P(\hat{Y}=1|Y=1,G=0) - P(\hat{Y}=1|Y=1,G=1)|$\\
$+ |P(\hat{Y}=1|Y=0,G=0) - P(\hat{Y}=1|Y=0,G=1)|$
} \\
\bottomrule
\end{tabular}%
}
\end{table}

Beyond global fidelity, fairness metrics show that a stolen model can match the victim's outputs on average while redistributing errors across demographic groups in harmful ways. An attacker optimizing only fidelity may land on a clone whose decisions differ precisely in regions or subpopulations where margins are small. Multiplicity metrics reveal where disagreements are possible, but fairness metrics tell us who bears them. %Small threshold or calibration shifts can leave fidelity unchanged yet amplify disparate impact, creating legal, ethical, and operational exposure.

\section{Experiments}
\label{sec:Exp}
 \begin{figure*}[h!]
    \begin{center}
    \includegraphics[width=0.95\textwidth]{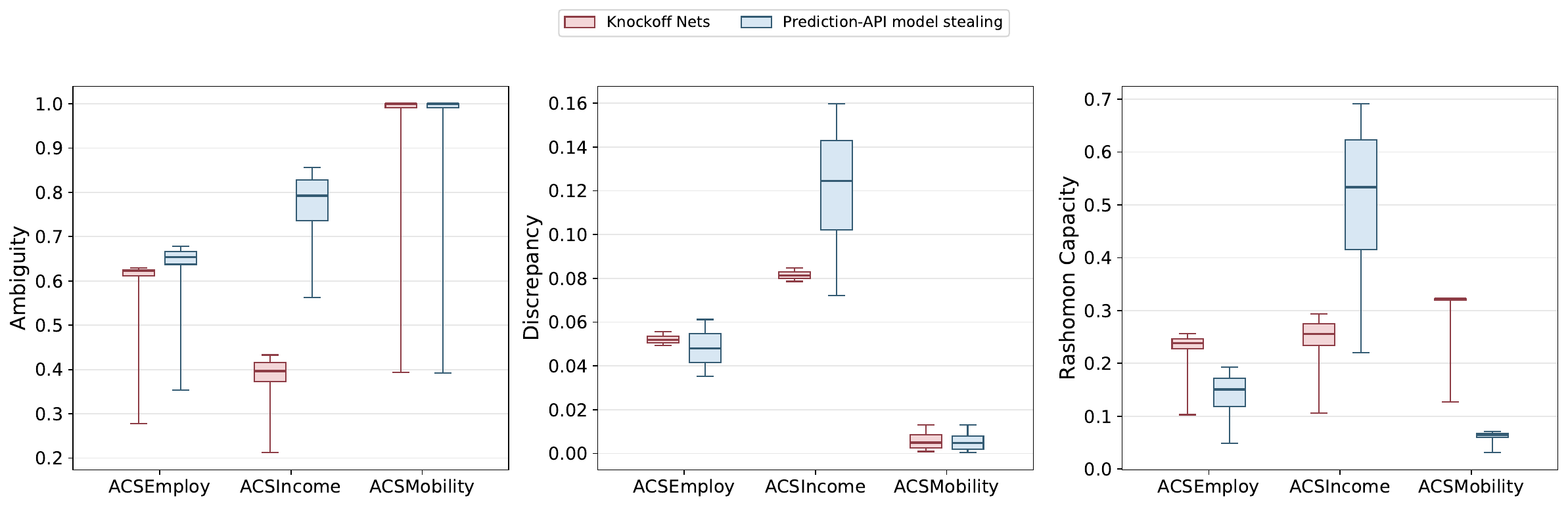}
    \end{center}
    \caption{Multiplicity results for dropout-based \rashomon{} sets on \folktables{} datasets. Each box summarizes the metric across dropout rates $p \in \{0.01,0.02,\ldots,0.10\}$, for each dropout rate, $5000$ stochastic surrogate models are sampled.}
    \label{fig:results_amulet}
\end{figure*}

 \begin{figure*}[h!]
    \begin{center}
    \includegraphics[width=0.95\textwidth]{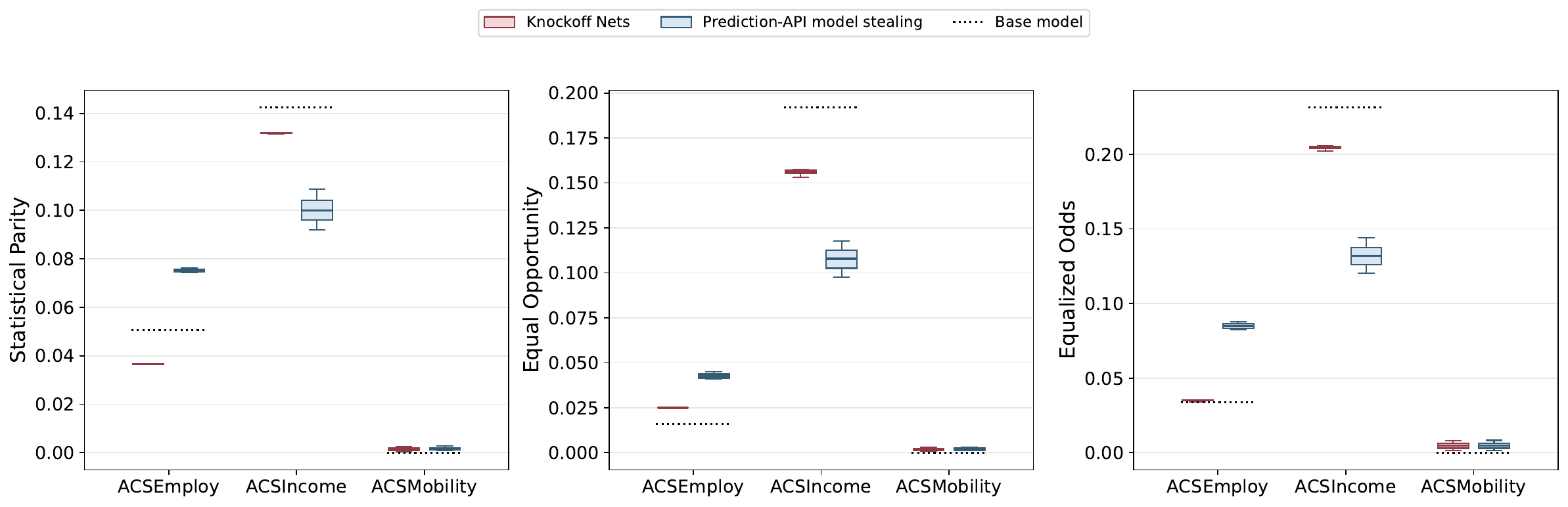}
    \end{center}
    \caption{Fairness metrics for \rashomon{} for both attacks. Each box summarizes the metric across dropout rates $p \in \{0.01,0.02,\ldots,0.10\}$, for each dropout rate, $5000$ stochastic surrogate models are sampled. (Predictive equality in Figure~\ref{fig:fairness_all}).}
    \label{fig:fairness_box}
\end{figure*}

We evaluate whether high-fidelity model extraction is sufficient to guarantee behavioral equivalence between a victim model and its stolen surrogate. Across all experiments, we ask if dropout-based \rashomon{} around extracted surrogates reveals predictive multiplicity hidden by aggregate fidelity. 

We study this question across three settings: tabular prediction, medical image classification, and NLP sentiment classification. This allows us to test whether extraction-induced multiplicity is specific to a particular architecture or data modality, or whether this phenomenon generalizes across different prediction tasks.

\subsection{Extraction pipeline}

All datasets are divided into partitions of $60\%$ training, $20\%$ attack (used to generate the surrogate model), and $20\%$ test data. 

For these experiments, we train one victim model and perform one extraction attack per dataset. The \rashomon{} is then generated by applying dropout to the extracted surrogate rather than by retraining multiple victim models. We use dropout values ranging from $0.01$ to $0.1$ in increments of $0.01$, and sample $5{,}000$ stochastic models for each dropout value, yielding $10$ dropout-based \rashomon{} sets of size $5{,}000$ per dataset.

We consider two complementary black-box extraction attacks, which will be reused across all subsequent experiments.

\paragraph{Knockoff Nets}
We use the Knockoff Nets extraction procedure~\citep{Orekondy_2019_CVPR}, which steals model functionality through black-box interactions alone. We use ART's (Adversarial Robustness Toolbox)~\citep{art2018} \texttt{PyTorchClassifier} interface for both networks and run Knockoff Nets with adaptive query sampling and a loss-based reward. To improve stability, the attack pool is balanced using the victim's predicted classes before extraction.

\paragraph{Prediction-API stealing}
We follow the standard black-box model stealing setting~\citep{tramer2016stealing,jagielski2020high}, in which an adversary queries the target model on an auxiliary dataset, collects the returned posterior outputs, and uses these outputs as supervision to train a stolen model that approximates the target model's behavior. We instantiate this setting using the model extraction pipeline provided by the \texttt{amuletml}~\citep{dudduSoKUnintendedInteractions2024} package.

\subsection{Experiment 1: Extraction of neural networks for tabular data}

\paragraph{Motivation}
We study whether predictive multiplicity occurs on tabular data when the surrogate is obtained via black-box extraction rather than being trained directly from labeled data, and whether this phenomenon holds across distinct extraction frameworks.
 \begin{figure*}[h!]
    \begin{center}
    \includegraphics[width=0.75\textwidth]{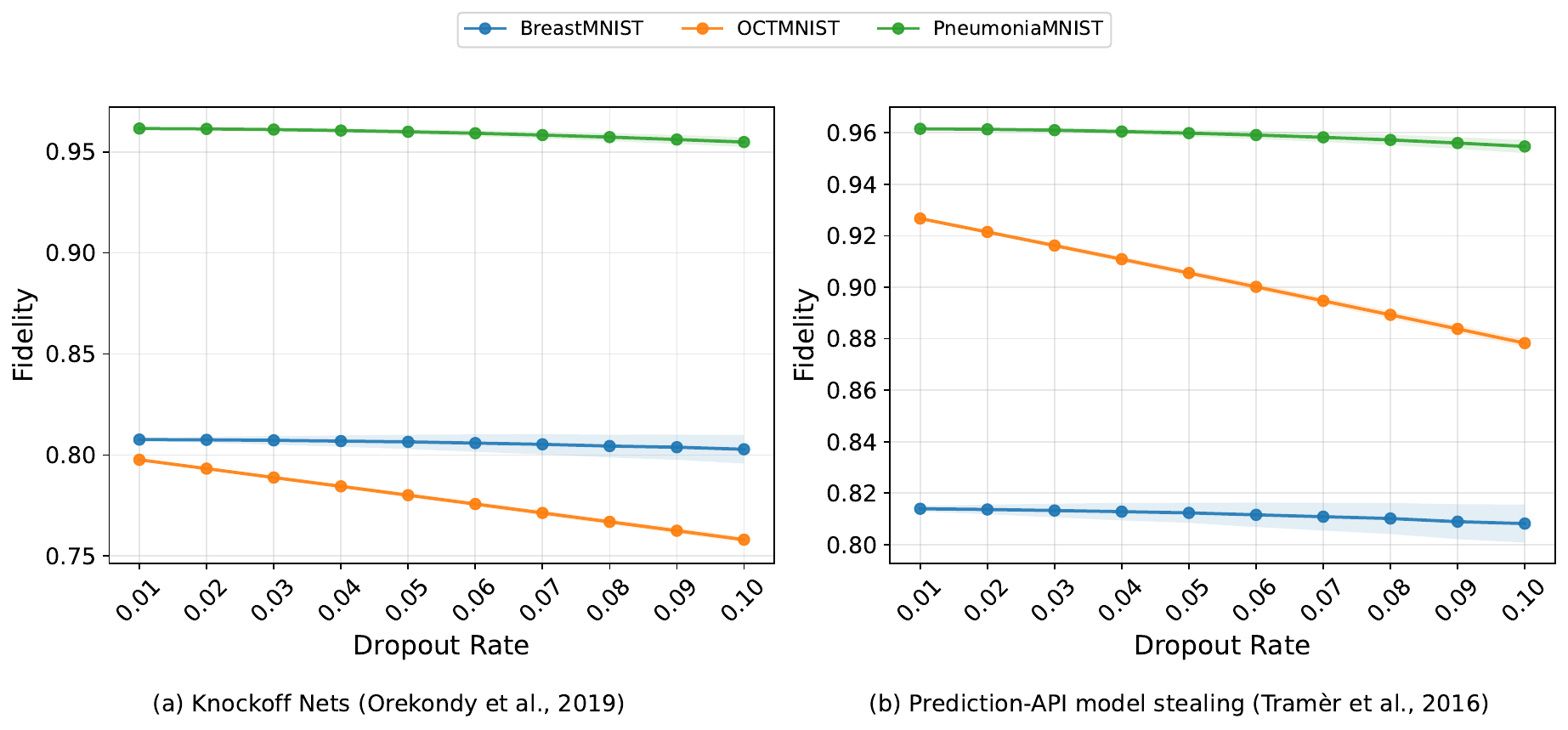}
    \end{center}
    \caption{Fidelity for \rashomon{} of size 5000 for each dropout rate for both attacks on \texttt{MedMNIST} datasets.}
    \label{fig:fid_image}
\end{figure*}

\paragraph{Datasets}
We conduct our experiments on three widely used benchmark datasets from \folktables{}~\citep{ding2021retiring} employed in the fair machine learning literature. All datasets are derived from the 1994 US Census survey. Those datasets are: \texttt{ACSIncome} (\emph{income superior to 50,000\$}), \texttt{ACSMobility} (\emph{home did not change since a year ago}) and \texttt{ACSEmploy} (\emph{employed}). To compute the fairness metrics, we define majority and minority groups using the race attribute \texttt{RAC1P}~\citep{ding2021retiring}, which is included among the predictive features in all three tasks. We use \textit{White alone} as the majority group and \textit{Black or African American alone} as the marginalized group.

\paragraph{Experimental setup}
For each dataset, both the victim and extracted surrogate are feed-forward neural networks with one hidden layer of $1{,}000$ ReLU units. 

\paragraph{Results}
Figure~\ref{fig:fid} first shows that the extracted surrogates operate in a high-fidelity regime across the three tabular tasks. Under Knockoff Nets, fidelity remains above $0.91$ for all datasets, with \texttt{ACSMobility} close to perfect agreement and \texttt{ACSEmployment} around $0.95$. Under prediction-API stealing, fidelity is also high for \texttt{ACSMobility} and \texttt{ACSEmployment}, remaining above $0.94$, while \texttt{ACSIncome} stays mostly above $0.85$ despite a more visible decline as the dropout rate increases. These results indicate that the extracted surrogates retain strong aggregate agreement with the victim model across attacks and datasets.

Figure~\ref{fig:results_amulet} shows the distribution of ambiguity, discrepancy, and Rashomon capacity across all dropout-based Rashomon sets. We observe high ambiguity across all datasets, ranging from $20\%$ to $100\%$. This means that between $20\%$ and $100\%$ of individuals receive conflicting predictions across models in the dropout-based Rashomon sets. For example, in the low-dropout regime ($\leq 0.03$), where fidelity is very high, ambiguity ranges from $0.28$ to $0.65$ for \texttt{ACSEmploy}, from $0.20$ to $0.74$ for \texttt{ACSIncome}, and from $0.40$ to $1.00$ for \texttt{ACSMobility} as highlighted by Figure~\ref{fig:amb}. Figure~\ref{fig:disc}, instead, shows that the discrepancy is more moderate in the low-dropout regime. For \texttt{ACSIncome}, we observe discrepancy values between $8\%$ and $10\%$. This means that, even among high-fidelity surrogates, one can find a competing surrogate that assigns conflicting predictions to up to $10\%$ of individuals. Discrepancy ranges from $0.038$ to $0.05$ for \texttt{ACSEmploy} and remains close to $0.0$ for \texttt{ACSMobility}. Rashomon capacity is largest for \texttt{ACSIncome} under prediction-API stealing, increasing from $0.2204$ to $0.6915$ (\emph{see} Figure~\ref{fig:rc}). It remains lower for \texttt{ACSMobility}, suggesting that \texttt{ACSIncome} exhibits stronger score-level variation across near-equivalent stolen models. Overall, these results show that high-fidelity extraction does not identify a unique behavioral clone of the victim model: many individuals may receive different predictions depending on which near-equivalent surrogate is selected, even when all surrogates preserve similar aggregate fidelity. Figure~\ref{fig:fairness_box} shows that this predictive multiplicity also affects group fairness estimates. Fairness remains relatively concentrated for \texttt{ACSMobility}, but varies more strongly for \texttt{ACSIncome} and \texttt{ACSEmploy}, including shifts in equal opportunity and statistical parity under prediction-API stealing.
%Figures~\ref{fig:fid} and~\ref{fig:results_amulet} show that black-box model stealing attacks can produce high-fidelity surrogates that still exhibit substantial predictive multiplicity on tabular data tasks. \texttt{ACSEmploy} is the most stable task: fidelity remains high under both Knockoff Nets ($0.9512$--$0.9455$) and prediction-API stealing ($0.9656$--$0.9408$), with discrepancy around $0.05$. In contrast, \texttt{ACSIncome} and \texttt{ACSMobility} exhibit much stronger multiplicity. Under prediction-API stealing on \texttt{ACSIncome}, ambiguity rises from $0.6808$ to $0.8557$ and Rashomon Capacity from $0.2204$ to $0.6915$; under Knockoff Nets on \texttt{ACSMobility}, ambiguity reaches $1.0$ and Rashomon Capacity remains around $0.3225$. Figure~\ref{fig:fairness_box} shows that this predictive multiplicity also affects group fairness estimates: fairness remains relatively concentrated for \texttt{ACSMobility}, but varies more strongly for \texttt{ACSIncome} and \texttt{ACSEmploy}, including shifts in equal opportunity and statistical parity under prediction-API stealing. Overall, extraction fidelity alone is insufficient to characterize the behavior or fairness profile of the stolen model.

 \begin{figure*}[h!]
    \begin{center}
    \includegraphics[width=0.95\textwidth]{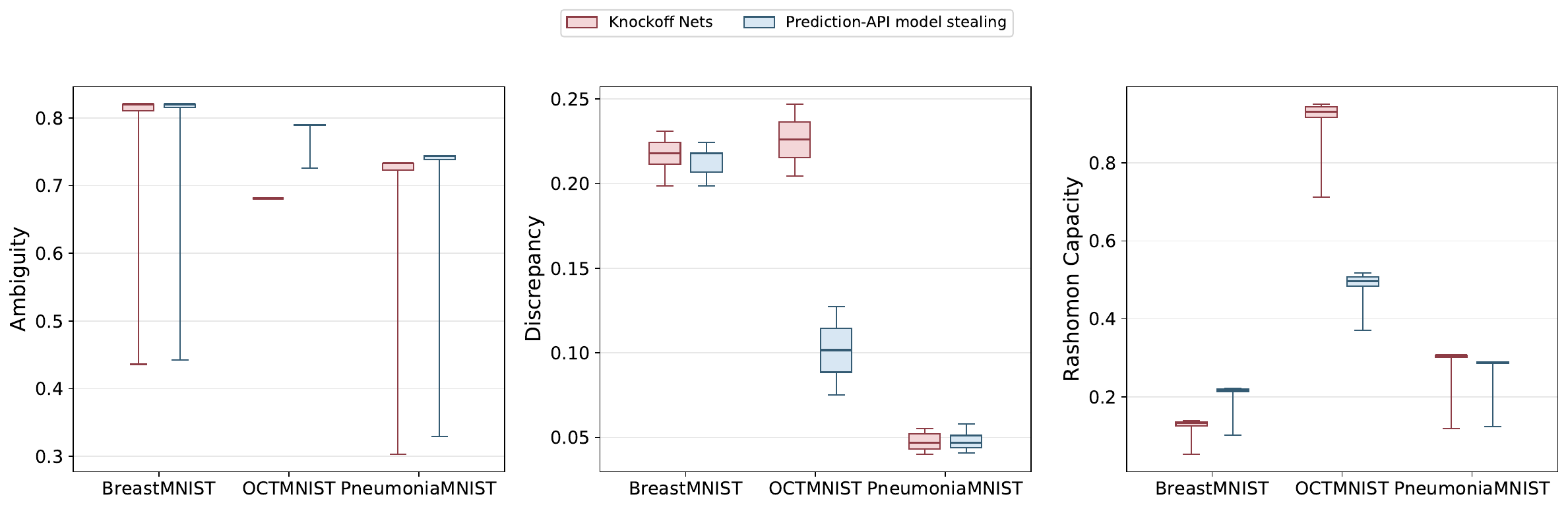}
    \end{center}
    \caption{Multiplicity results for \rashomon{} for both attacks on \texttt{MedMNIST} datasets. Each box summarizes the metric across dropout rates $p \in \{0.01,0.02,\ldots,0.10\}$, for each dropout rate, $5000$ stochastic surrogate models are sampled.}
    \label{fig:box_metrics_image}
\end{figure*}

\subsection{Experiment 2: Extraction of neural networks for medical imaging and NLP}

\paragraph{Motivation}
We next examine whether the predictive multiplicity observed after extracting tabular models also appears in high-dimensional domains. Medical imaging and NLP provide two complementary settings: the former uses deep convolutional models over visual inputs, while the latter uses transformer-based language models over text. Together, these experiments test whether dropout-based \rashomon{} remain informative after black-box extraction beyond tabular neural networks.

\paragraph{Datasets}
For medical imaging, we use three datasets from \texttt{MedMNIST}~\citep{yangMedMNISTV2Largescale2023}: \texttt{BreastMNIST} (2 classes), \texttt{OCTMNIST} (4 classes), and \texttt{PneumoniaMNIST} (2 classes), covering breast ultrasound, optical coherence tomography, and chest X-ray classification. For NLP, we use \texttt{FST}, a binary financial sentiment classification task derived from \texttt{financial phrase-bank}~\citep{maloGoodDebtBad2014}. 

\paragraph{Experimental setup}
For the medical imaging experiments, both the victim and extracted surrogate are ResNet-50 models~\citep{he2016deep} initialized with ImageNet-pretrained weights. For the NLP experiment, both the victim and surrogate are \texttt{BERT} sequence classifiers~\citep{DBLP:journals/corr/abs-1810-04805}. Because the NLP experiments are more computationally expensive, we use the same dropout grid from $0.01$ to $0.10$ but sample $500$ epsilon-filtered stochastic models per dropout value. Dropout is applied throughout the extracted BERT surrogate by perturbing all linear layers.

\paragraph{Medical imaging results}
Figure~\ref{fig:fid_image} highlights that \texttt{PneumoniaMNIST} is the clearest high-fidelity case: both attacks remain close to $0.96$ fidelity across the entire dropout range, not only for small dropout values. The extracted surrogate appears highly faithful under the usual agreement metric. Nevertheless, Figure~\ref{fig:box_metrics_image} shows that ambiguity rises from $0.3029$ to $0.7201$ under Knockoff Nets and from $0.3294$ to $0.7372$ under prediction-API stealing, while discrepancy remains low, below $0.0435$ (\emph{see} Figure~\ref{fig:disc_image}).

Under prediction-API stealing for \texttt{OCTMNIST}, fidelity is high at low dropout, ranging from $0.9267$ to $0.9162$ for dropout values up to $0.03$. Yet, the corresponding \rashomon{} already presents high ambiguity ($0.7259$--$0.7891$) and substantial Rashomon Capacity ($0.3713$--$0.4820$) in the low-dropout regime as shown on Figures~\ref{fig:amb_image}--\ref{fig:rc_image}. This indicates that the sampled surrogates do not merely flip a small number of isolated labels: their predictive scores vary substantially across the \rashomon{}, so the extracted model leaves a broad local region of plausible probability assignments.

Finally, for \texttt{BreastMNIST}, fidelity is around $0.81$ for both attacks, yet ambiguity still reaches above $0.80$ by dropout $0.03$. Overall, the image experiments show that high fidelity, when it is achieved, can coexist with substantial local multiplicity, and that different attacks can produce different \rashomon{}.

\begin{figure*}[h!]
    \begin{center}
    \includegraphics[width=0.95\textwidth]{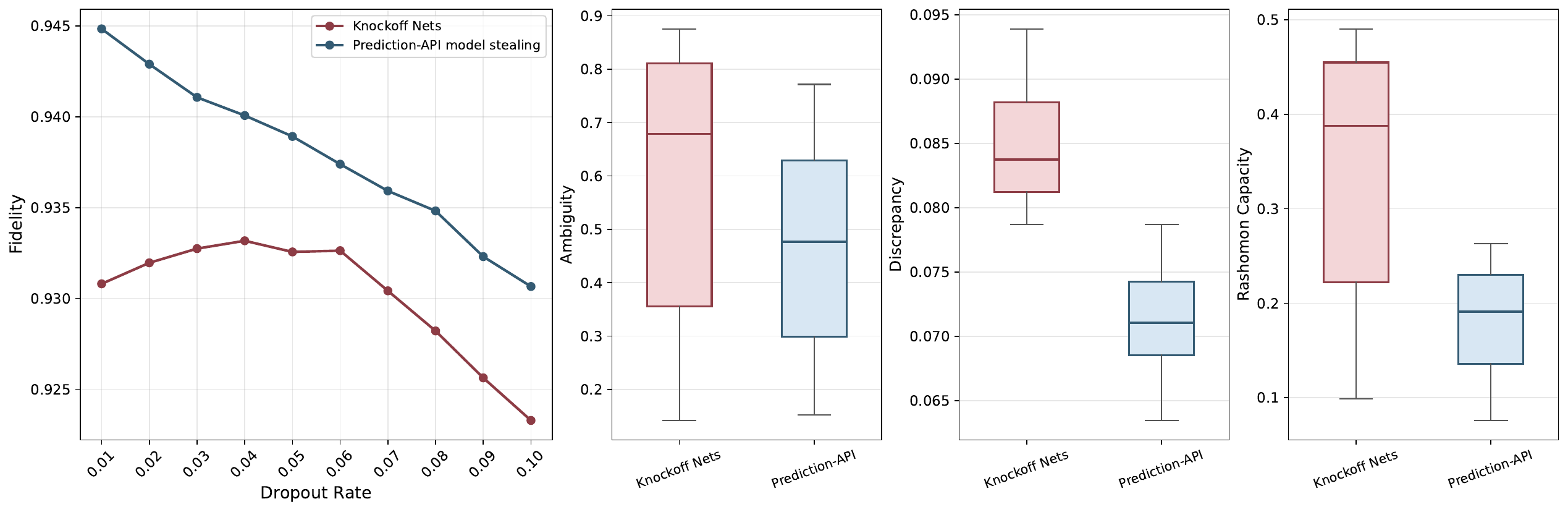}
    \end{center}
    \caption{Results for \rashomon{} for both attacks on FST dataset. Each box summarizes the metric across dropout rates $p \in \{0.01,0.02,\ldots,0.10\}$, for each dropout rate, $500$ stochastic surrogate models are sampled.}
    \label{fig:nlp}
\end{figure*}

\paragraph{NLP results}
Figure~\ref{fig:nlp} shows that the extracted \texttt{BERT} surrogates remain highly faithful to the victim on \texttt{FST}. Under Knockoff Nets, fidelity stays between $0.9308$ and $0.9233$ across the dropout range, while prediction-API stealing remains slightly higher, between $0.9448$ and $0.9307$. Thus, unlike settings where multiplicity could be attributed to poor extraction, the NLP experiment remains firmly in a high-fidelity regime for both attacks.

The multiplicity metrics reveal a different behavior: ambiguity grows rapidly even while fidelity changes only mildly. In the low-dropout regime ($\leq 0.03$), Figure~\ref{fig:nlp_all} shows that ambiguity already increases from $0.1421$ to $0.3223$ under Knockoff Nets and from $0.1523$ to $0.3223$ under prediction-API stealing.

Rashomon Capacity follows the same gradual augmentation. In the low-dropout regime, it rises from $0.0986$ to $0.2037$ under Knockoff Nets and from $0.0758$ to $0.1436$ under prediction-API stealing (Figure~\ref{fig:nlp_all}). At higher dropout values, this effect becomes much stronger, with ambiguity reaching $0.8680$ for Knockoff Nets and $0.7716$ for prediction-API stealing. Overall, the NLP experiment shows that even a consistently high-fidelity extracted language model can contain a progressively expanding \rashomon{} of alternative predictions.

\section{Discussion}

Across tabular data, medical imaging, and NLP, our experiments show that the main risk in model extraction is not captured by fidelity alone. A stolen model can agree with the victim on aggregate while still belonging to a \rashomon{} set with substantial predictive variation. This is visible even in high-fidelity regimes: \texttt{PneumoniaMNIST} and \texttt{FST} preserve strong agreement with the victim, yet still exhibit large ambiguity under dropout sampling. Thus, extraction success should not be interpreted as behavioral equivalence. For downstream use, auditing, or attack transfer, a surrogate that appears faithful globally may still differ from the victim on specific inputs.

The structure of this multiplicity depends on both the dataset and the extraction pipeline. Knockoff Nets and prediction-API stealing do not merely produce different fidelity values; they can induce different local neighborhoods around the extracted surrogate. In some cases, such as \texttt{OCTMNIST}, lower fidelity is accompanied by much larger Rashomon Capacity, while in others, high-fidelity extraction still leaves substantial ambiguity. This suggests that two attacks with similar aggregate performance may yield surrogate families that disagree in different ways and on different examples. 

The tabular experiments further show that predictive multiplicity can affect fairness performances. Even when a surrogate appears to approximate the victim well, the fairness profile of models sampled around that surrogate can vary, especially for \texttt{ACSIncome} and \texttt{ACSMobility}. This suggests that claims of faithful copying should be evaluated not only through aggregate fidelity, but also through whether the surrogate preserves the victim's group-level performance patterns across populations.

This work has several limitations. First, our \rashomon{} sets are approximated through dropout-based sampling rather than exhaustive enumeration, so the reported variability captures a local neighborhood around the extracted surrogate. However, because our approximate \rashomon{} sets already reveal substantial multiplicity, they are sufficient to demonstrate a concrete risk; the approximation may underestimate the full space of near-equivalent surrogates, but the observed variability is already large enough to challenge claims of faithful copying. Second, each domain uses a fixed architecture family: neural networks for tabular data, ResNet-50 for medical imaging, and BERT for NLP. This corresponds to a strong-adversary setting in which the surrogate architecture is well aligned with the victim architecture. Relaxing this assumption by considering heterogeneous victim and surrogate architectures would likely enlarge the space of near-equivalent surrogates and may induce even greater multiplicity. Such experiments would also provide a broader picture of the extent to which supposedly faithful copies can diverge in deployment-relevant performance across populations. Overall, our results should therefore be viewed as a conservative estimate of extraction-induced multiplicity. Even under favorable conditions for producing a faithful copy, high fidelity does not eliminate substantial variation in downstream behavior. Finally, while we focus on predictive multiplicity and tabular fairness, other behavioral notions remain open, including explanation agreement, robustness, calibration, and adversarial transferability. 

\section{Conclusion}
In this paper, we revisited the standard interpretation of model stealing success by showing that high fidelity does not, by itself, imply that a stolen model is practically interchangeable with its victim. By studying extraction through the lens of the \texttt{RASHOMON} effect, we showed that query-based supervision can leave the surrogate underdetermined, yielding families of near-equivalent extracted models rather than a unique faithful clone. Across Knockoff Nets and prediction-API model stealing, and across tabular, medical imaging, and NLP tasks, extracted surrogates exhibited substantial multiplicity even when fidelity remained high.

Our results show that this multiplicity appears at several levels. At the prediction level, extracted \rashomon{} can contain models with large ambiguity, discrepancy, and Rashomon Capacity. At the fairness level, tabular surrogates with similar fidelity can induce different group-level outcomes.

These findings suggest that fidelity should be viewed as only one dimension of extraction quality, rather than as a sufficient proxy for functional equivalence. The practical consequences of model stealing depend not only on how closely a surrogate matches the victim on average, but also on the structure of the extraction-induced \rashomon{} surrounding it. This has direct implications for attack evaluation and defense design in \texttt{MLaaS} settings: assessing model extraction risk requires measuring predictive multiplicity, and when relevant fairness variability, alongside conventional agreement-based metrics. More broadly, our work places model stealing within the wider literature on multiplicity by showing that extraction itself constitutes a source of non-uniqueness at deployment time.

\section*{Acknowledgements}
The authors thank the Digital Research Alliance of Canada for computing resources. Ulrich A\"ivodji is supported by the Fonds de recherche du Qu\'ebec -- Nature et technologies (FRQNT) Research Support for New Academic grant (342675).

\bibliographystyle{plainnat}
\bibliography{references}

\clearpage
\appendix
\section*{Appendix}
\label{app:additional_results}

This appendix provides complementary plots that decompose the aggregate results reported in Section~\ref{sec:Exp}. 

\begin{figure}[h!]
    \centering
    \includegraphics[width=0.7\textwidth]{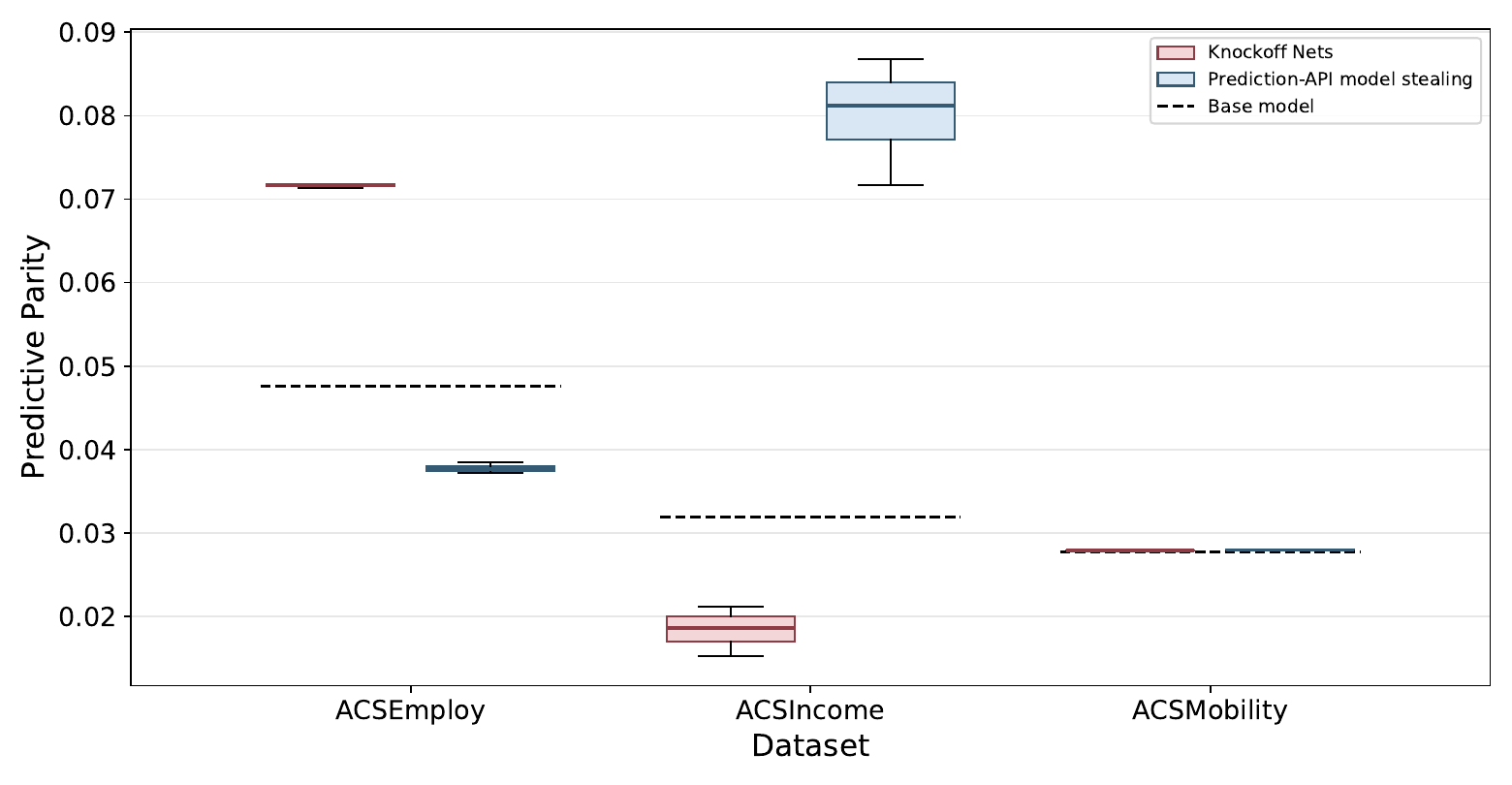}
    \caption{Predictive Equality for \rashomon{} of size 5000 for both attacks.}
    \label{fig:fairness_all}
\end{figure}

\paragraph{Additional fairness results.}
Figure~\ref{fig:fairness_all} reports predictive equality for the tabular experiments. It complements the fairness results in Figure~\ref{fig:fairness_box} by showing how another group fairness metric varies across the dropout-based \rashomon{}.

\begin{figure*}[h!]
    \begin{center}
    \includegraphics[width=0.7\textwidth]{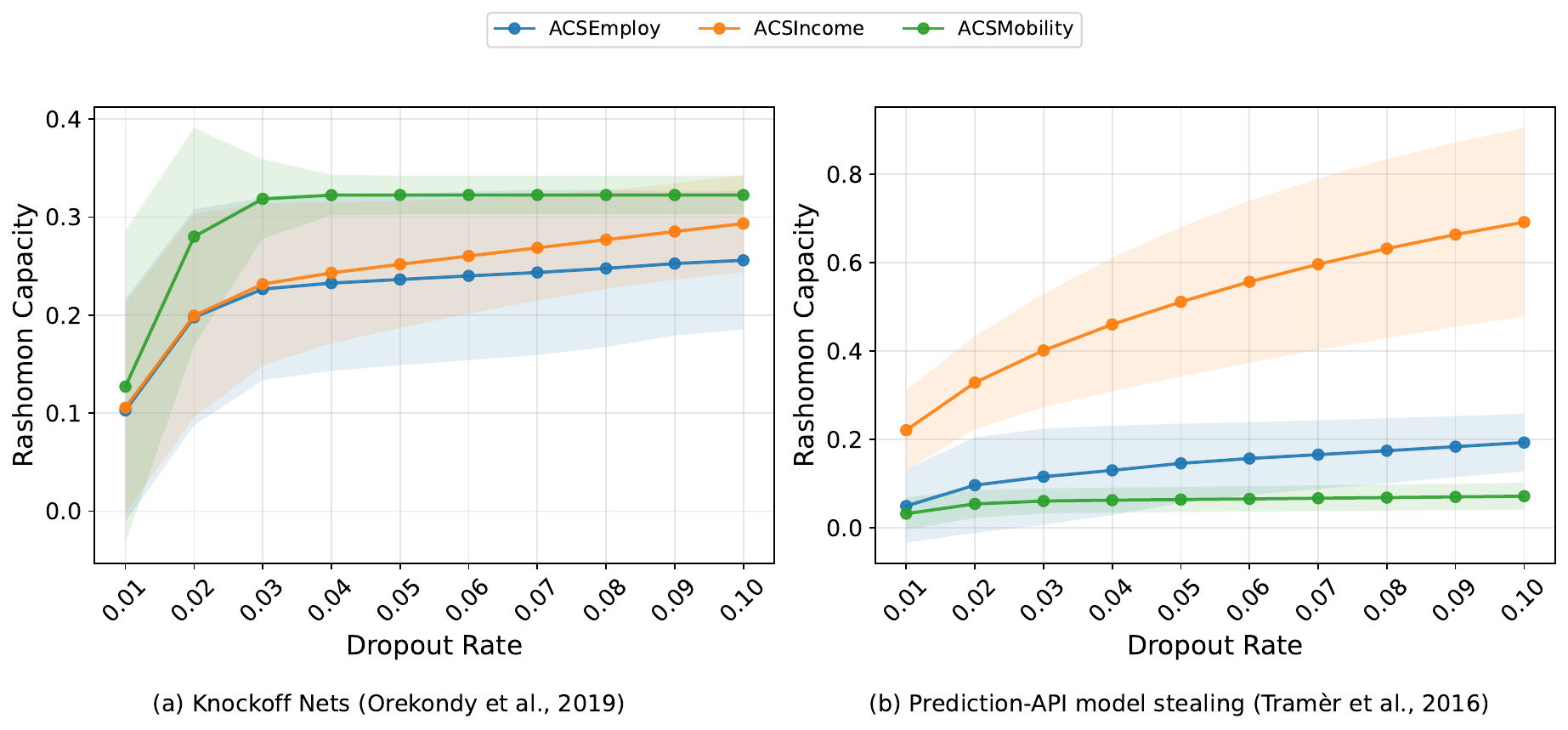}
    \end{center}
    \caption{Rashomon Capacity for \rashomon{} of size 5000 for each dropout rate for both attacks on \folktables{} datasets.}
    \label{fig:rc}
\end{figure*}

\paragraph{Tabular multiplicity results.}
Figures~\ref{fig:amb}, \ref{fig:disc}, and \ref{fig:rc} provide the individual ambiguity, discrepancy, and Rashomon Capacity results for the \folktables{} datasets. These plots expand the summary shown in Figure~\ref{fig:results_amulet} and show how each metric evolves across dropout rates and extraction attacks.

\begin{center}
    \includegraphics[width=0.7\textwidth]{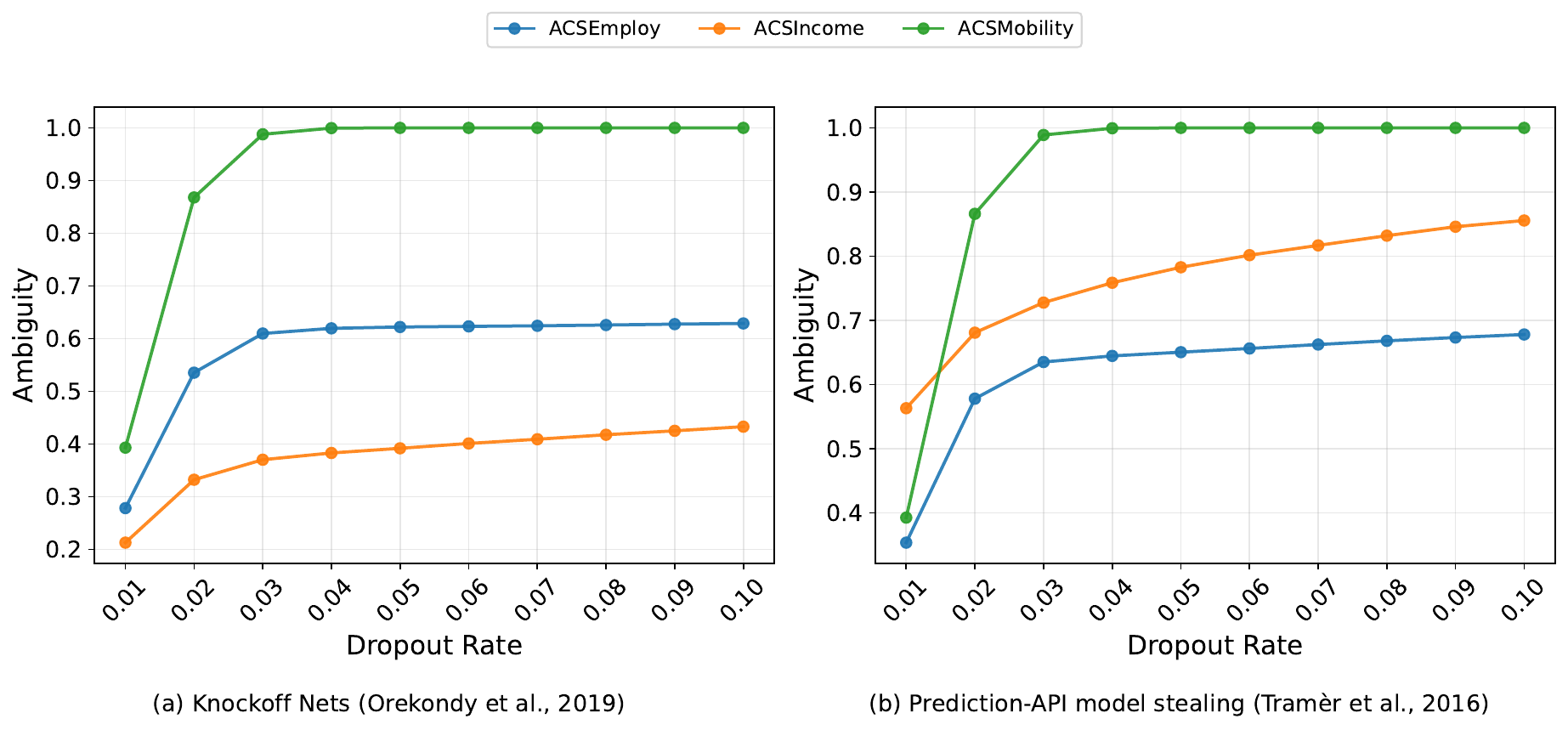}
    \captionof{figure}{Ambiguity for \rashomon{} of size 5000 for each dropout rate for both attacks on \folktables{} datasets.}
    \label{fig:amb}
\end{center}
\begin{center}
    \includegraphics[width=0.7\textwidth]{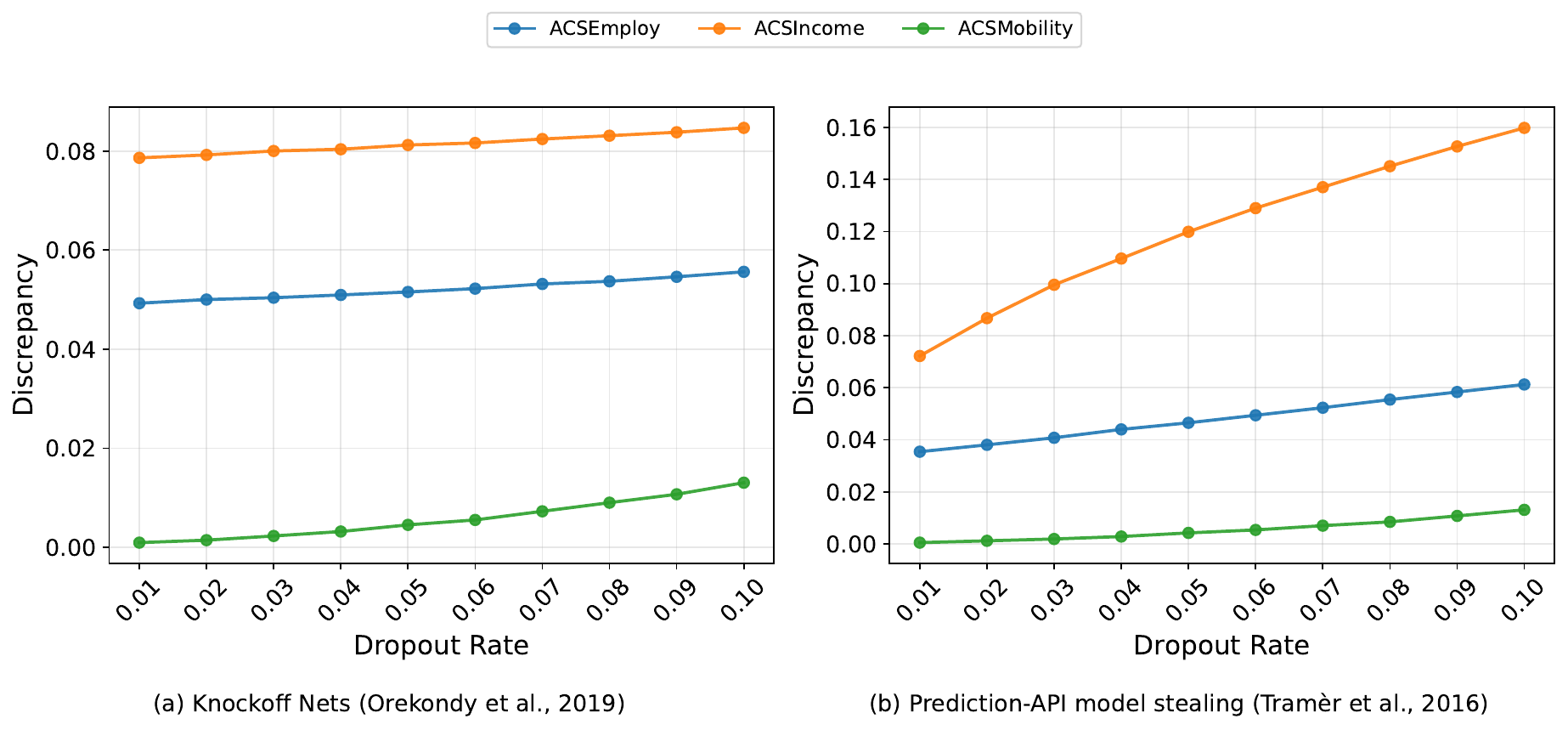}
    \captionof{figure}{Discrepancy for \rashomon{} of size 5000 for each dropout rate for both attacks on \folktables{} datasets.}
    \label{fig:disc}
\end{center}

\paragraph{Medical imaging multiplicity results.}
Figures~\ref{fig:amb_image}, \ref{fig:disc_image}, and \ref{fig:rc_image} report the corresponding multiplicity metrics for the \texttt{MedMNIST} experiments. They provide a more detailed view of the image-domain results summarized in Figure~\ref{fig:box_metrics_image}.

\begin{center}
    \includegraphics[width=0.7\textwidth]{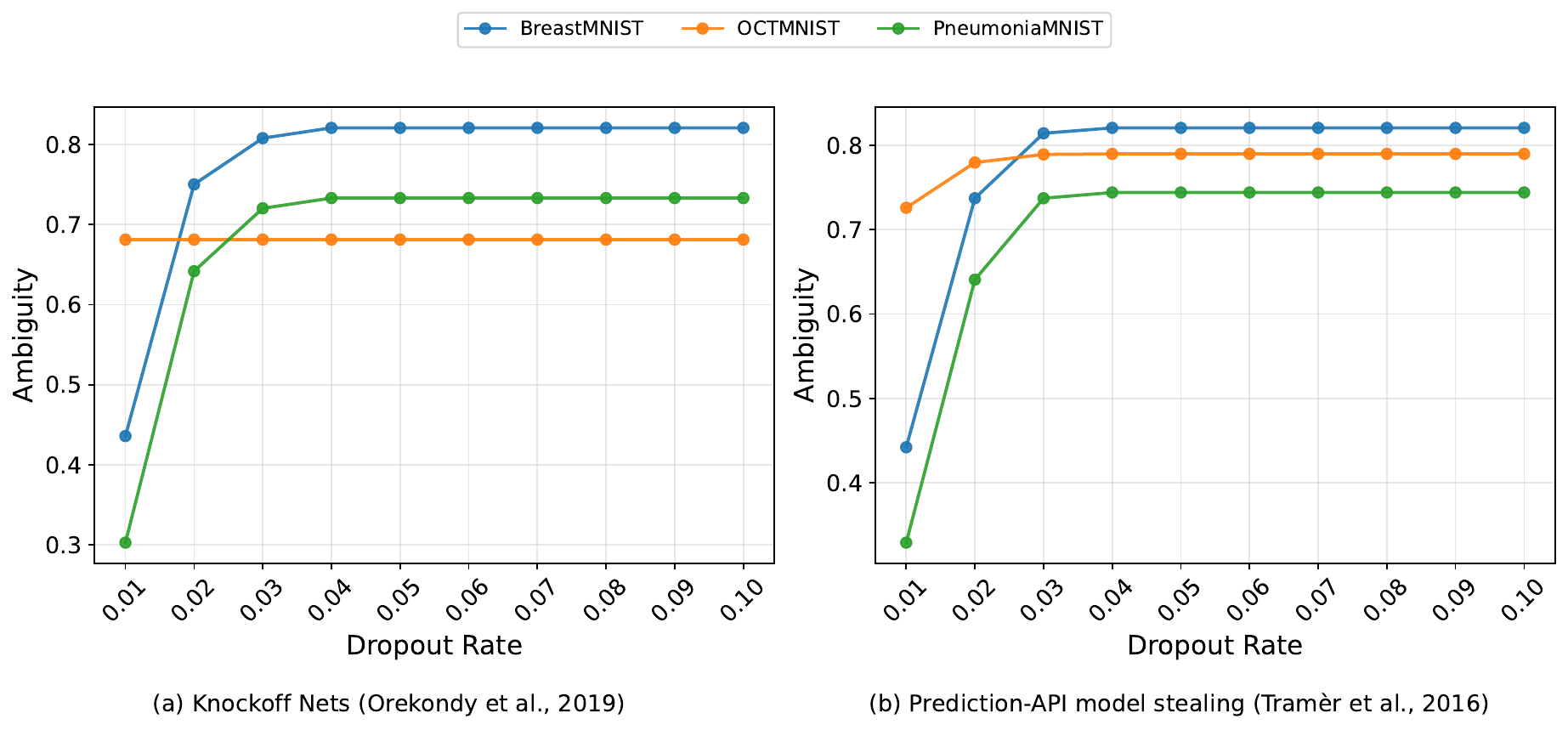}
    \captionof{figure}{Ambiguity for \rashomon{} of size 5000 for each dropout rate for both attacks on \texttt{MedMNIST} datasets.}
    \label{fig:amb_image}
\end{center}
\begin{center}
    \includegraphics[width=0.7\textwidth]{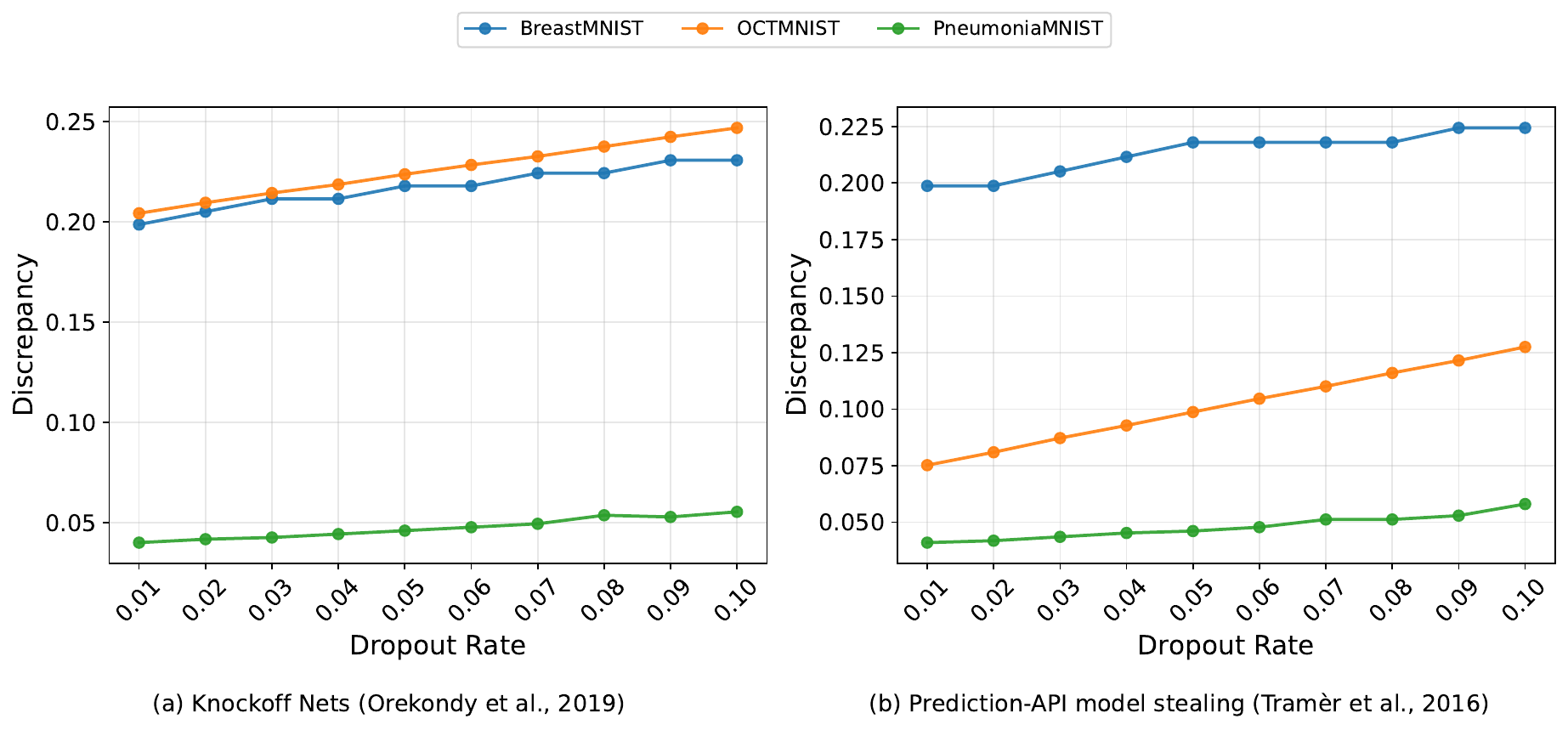}
    \captionof{figure}{Discrepancy for \rashomon{} of size 5000 for each dropout rate for both attacks on \texttt{MedMNIST} datasets.}
    \label{fig:disc_image}
\end{center}
\begin{center}
    \includegraphics[width=0.7\textwidth]{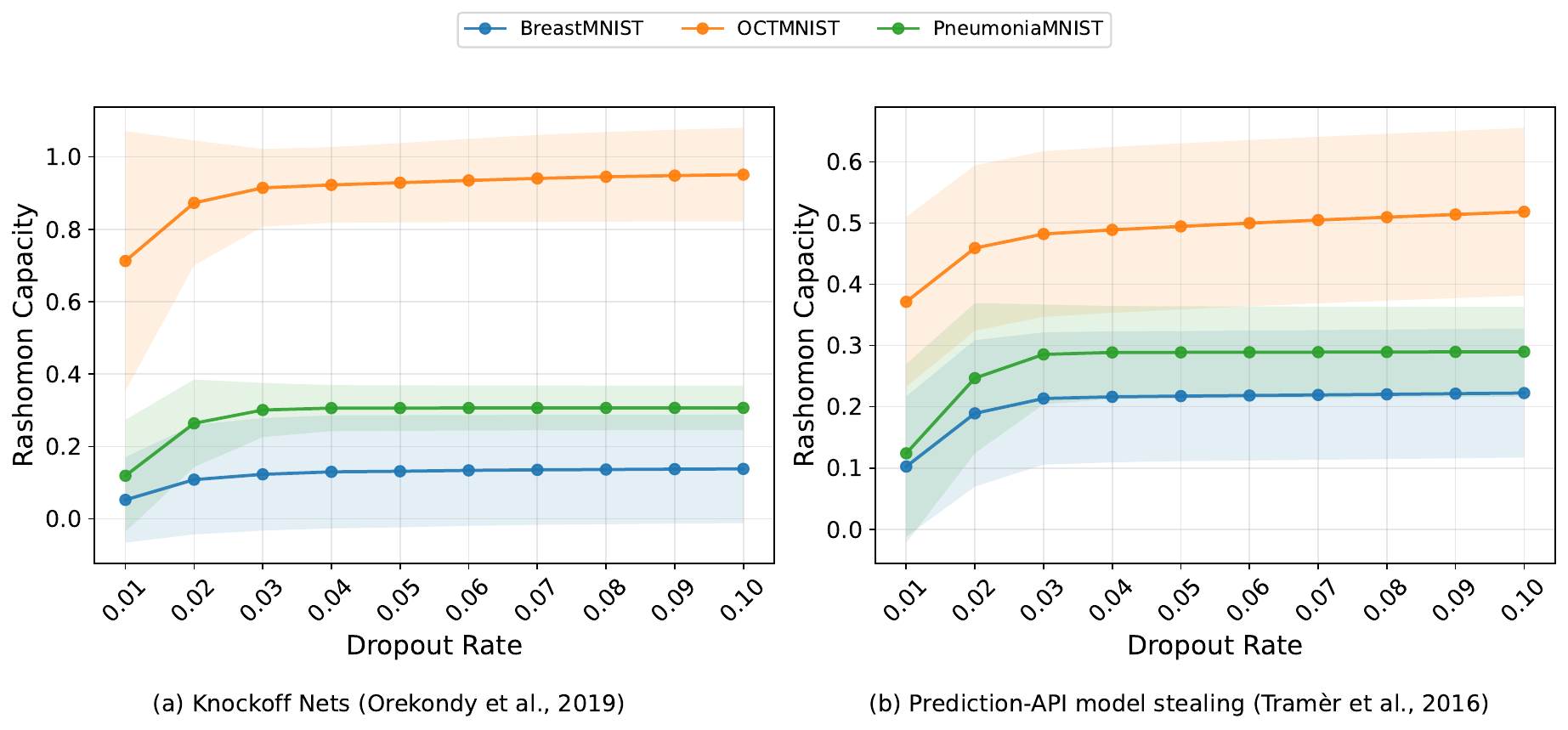}
    \captionof{figure}{Rashomon Capacity for \rashomon{} of size 5000 for each dropout rate for both attacks on \texttt{MedMNIST} datasets.}
    \label{fig:rc_image}
\end{center}

\begin{figure*}[h!]
    \centering
    \includegraphics[width=1\textwidth]{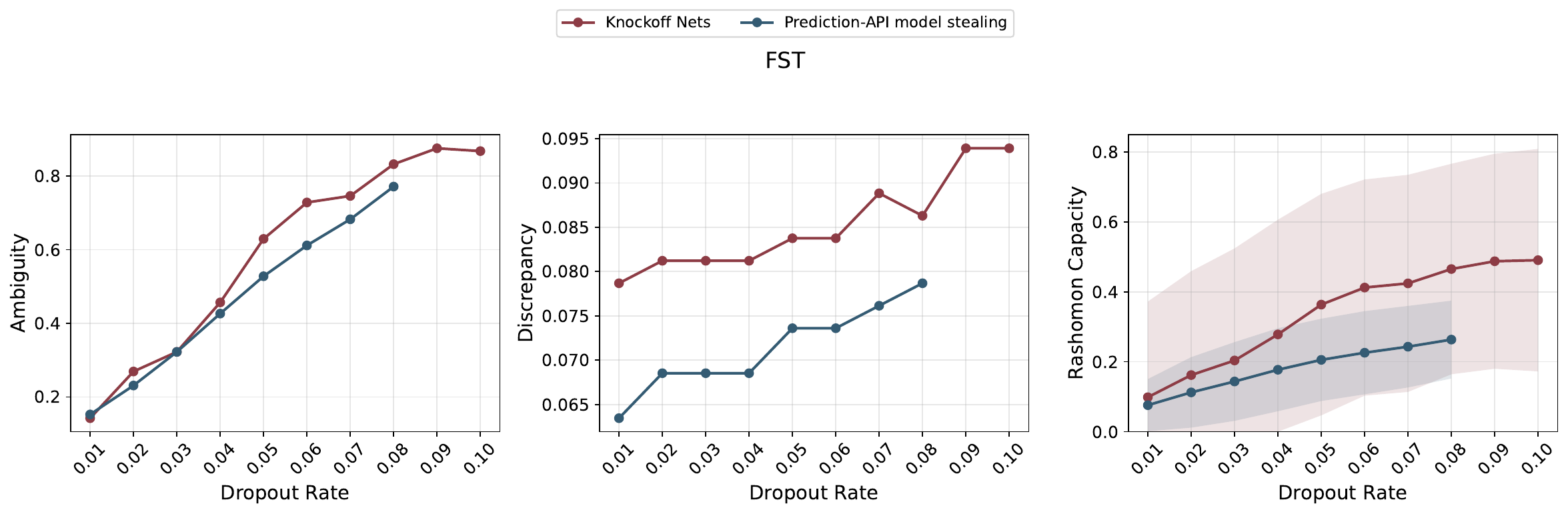}
    \caption{Ambiguity, Discrepancy and Rashomon Capacity for \rashomon{} of size 500 for each dropout rate for both attacks on \texttt{FST} dataset.}
    \label{fig:nlp_all}
\end{figure*}

\paragraph{NLP multiplicity results.}
Figure~\ref{fig:nlp_all} reports ambiguity, discrepancy, and Rashomon Capacity for the \texttt{FST} sentiment classification experiment. These results complement Figure~\ref{fig:nlp} by showing the individual multiplicity trends behind the aggregate NLP results.

\end{document}